\documentclass[letterpaper, 10 pt, journal, twoside]{IEEEtran}

\IEEEoverridecommandlockouts                              

\usepackage{amsmath}
\usepackage{mathtools,amssymb}
\usepackage{hyperref}
\usepackage{bm}
\usepackage{color}
\usepackage{xcolor}
\usepackage{cite}
\usepackage{booktabs}
\usepackage{multirow}
\usepackage[linesnumbered,ruled,vlined]{algorithm2e}
\usepackage[binary-units]{siunitx}
\usepackage{subcaption}
\usepackage{makecell}
\usepackage{graphicx}
\usepackage{balance}

\DeclareMathOperator*{\argmin}{arg\,min}

\usepackage{mwe,tikz}

\newcommand{\cspace}[0]{\mathcal{C}}
\newcommand{\world}[0]{\mathcal{W}}
\newcommand{\Cobs}[0]{\mathcal{C}_{obs}}
\newcommand{\Cfree}[0]{\mathcal{C}_{free}}

\newcommand{\Rthree}[0]{\mathbb{R}^3}
\newcommand{\SOthree}[0]{\mathbb{SO}(3)}
\newcommand{\obstacles}[0]{\mathcal{O}}
\newcommand{\robot}[0]{\mathcal{A}}

\newcommand{\disable}[1]{}
\newcommand{\norm}[1]{\left\lVert#1\right\rVert}

\DeclareMathOperator*{\minimize}{\text{minimize}}
\DeclareMathOperator*{\subjectto}{\text{s.t.}}

\title{Learning Minimum-Time Flight\\ in Cluttered Environments}

\author{Robert Penicka,  Yunlong Song, Elia Kaufmann, Davide Scaramuzza
\thanks{
Manuscript received: February, 24, 2022; Revised May, 20, 2022; Accepted May, 31, 2022.
This paper was recommended for publication by Editor Tetsuya Ogata upon evaluation of the Associate Editor and Reviewers' comments.
The authors are with the Robotics and Perception Group, Department of Informatics, University of Zurich, and Department of Neuroinformatics, University of Zurich and ETH Zurich, Switzerland (\protect\url{https://rpg.ifi.uzh.ch}). This work was supported as a part of NCCR Robotics, a National Centre of Competence in Research, funded by the Swiss National Science Foundation (grant number 51NF40\_185543), the European Union's Horizon 2020 Research and Innovation Programme under grant agreement No. 871479 (AERIAL-CORE) and the European Research Council (ERC) under grant agreement No. 864042 (AGILEFLIGHT).
Digital Object Identifier (DOI): 10.1109/LRA.2022.3181755}
}

\renewcommand\subsubsection[1]{\vspace{0pt}\noindent\textbf{#1.}}

\begin{document}

\setlength{\abovedisplayskip}{6pt}
\setlength{\belowdisplayskip}{6pt}
\setlength{\abovedisplayshortskip}{4pt}
\setlength{\belowdisplayshortskip}{4pt}

\maketitle

\begin{abstract}

We tackle the problem of minimum-time flight for a quadrotor through a sequence of waypoints in the presence of obstacles while exploiting the full quadrotor dynamics.
Early works relied on simplified dynamics or polynomial trajectory representations that did not exploit the full actuator potential of the quadrotor, and, thus, resulted in suboptimal solutions.
Recent works can plan minimum-time trajectories; 
yet, the trajectories are executed with control methods that do not account for obstacles. 
Thus, a successful execution of such trajectories is prone to errors due to model mismatch and in-flight disturbances.
To this end, we leverage deep reinforcement learning and classical topological path planning to train robust neural-network controllers for minimum-time quadrotor flight in cluttered environments.
The resulting neural network controller demonstrates substantially better performance of up to 19\% over state-of-the-art methods.
More importantly, the learned policy solves the planning and control problem simultaneously online to account for disturbances, thus achieving much higher robustness.
As such, the presented method achieves 100\% success rate of flying minimum-time policies without collision, while traditional planning and control approaches achieve only 40\%.
The proposed method is validated in both simulation and the real world, with quadrotor speeds of up to \SI{42}{\kilo\meter\per\hour} and accelerations of~3.6g.
\end{abstract}

\begin{IEEEkeywords}
Integrated Planning and Learning, Motion and Path Planning, Reinforcement Learning
\end{IEEEkeywords}

\vspace{-1.3em}
\section*{Supplementary Material}
{\small
\vspace{-0.3em}
\noindent \textbf{Video:} \url{https://youtu.be/wR1niZvI3pI}
\vspace{-0.7em}
}

\section{Introduction}




\IEEEPARstart{Q}{uadrotors} are among the most agile and maneuverable flying machines~\cite{spectrum2020acrobatics} and have recently shown a substantial increase in autonomy capabilities~\cite{Loquercio21FlightWild}.
This renders quadrotors the ideal platform for first responders to search for survivors as quickly as possible after natural disasters like earthquakes, forest fires, or floods. 
Though the astonishing agility of autonomous quadrotors has been demonstrated in many research labs~\cite{mellinger2011minimum, foehn2020cpc,Loquercio21FlightWild,zhou2020raptor,Loquercio20DeepDroneRacing,kaufmann2020RSS,Kartik18FastFlight,Ryou21Blackbox,han2021fastracing}, planning minimum-time trajectories in cluttered environments and navigating them without collision remains an open problem.
%
To push research in the field of agile navigation and minimum-time planning, autonomous drone racing has emerged as a research field, with international competitions being organized, such as the Autonomous Drone Racing series at the recent IROS and NeurIPS conferences~\cite{moon2019challenges,guerra2019flightgoggles,Madaan20arxiv} and the AlphaPilot challenge~\cite{foehn2020alphapilot,han2021fastracing}.
Drone racing requires flying a drone through a sequence of gates or doorways in minimum time while avoiding collisions with the environment,
which is an ideal benchmark scenario for autonomous quadrotors being deployed in search and rescue scenarios.

\begin{figure}[t]
   \centering
   \includegraphics[width=1.0\columnwidth,height=0.57\columnwidth,trim=0 0 0 20, clip]{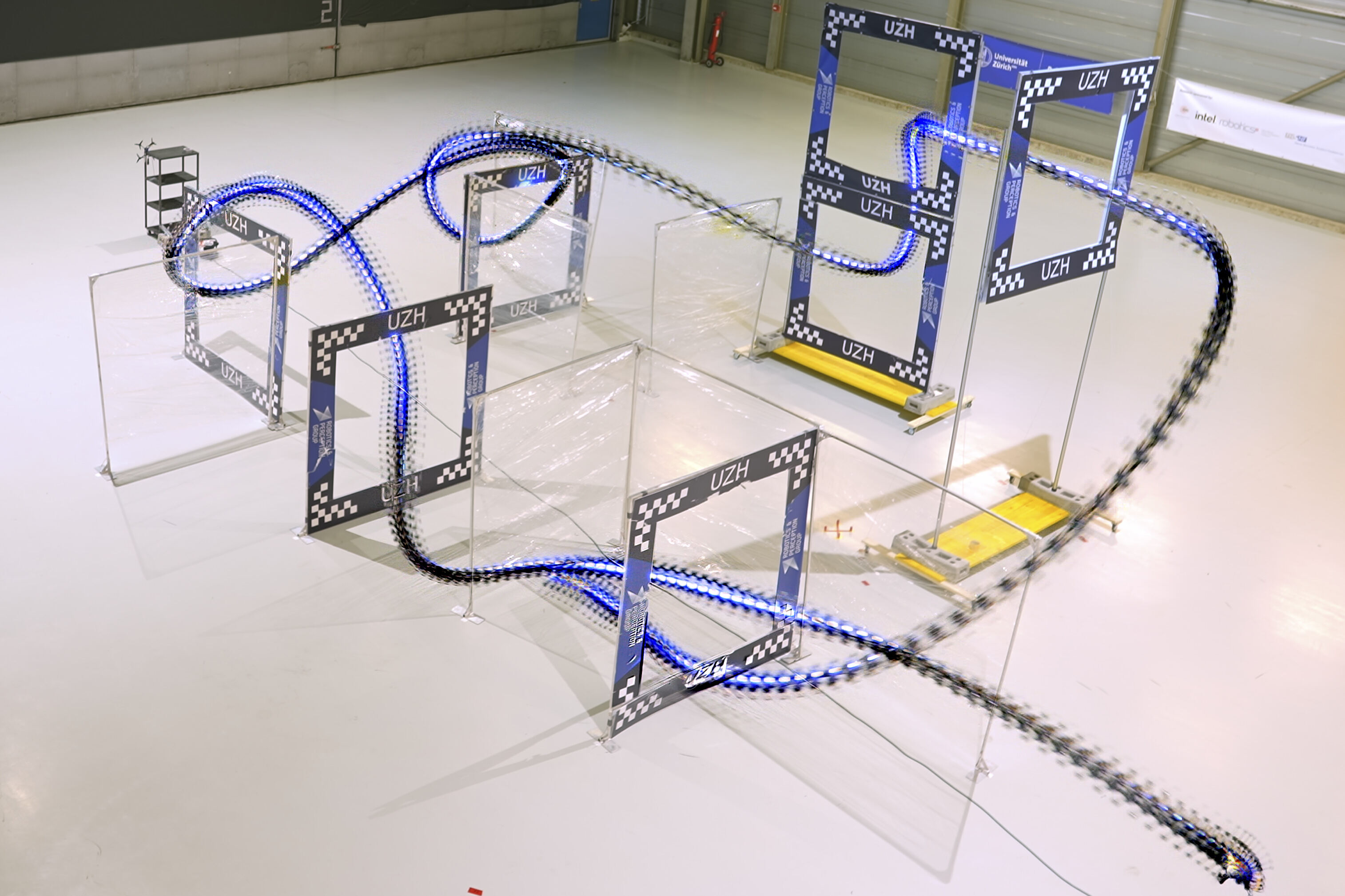}
   \vspace{-1.6em}
   \caption{\label{fig:illustration}
    Our quadrotor races through a complex race track in the real world while avoiding a set of obstacles made from transparent foil. 
    While racing, the quadrotor reaches speeds of up to \SI{42}{\kilo\meter\per\hour} and accelerates with up to~3.6g.
    \vspace{-1.8em}
   }
\end{figure}

%
%
%
By definition, the minimum-time objective requires the navigation and planning algorithm to constantly push the platform to its limits and operate the vehicle near the boundary of its physical envelope. 
Furthermore, the presence of 3D obstacles results in a highly nonconvex optimization problem that quickly becomes intractable to solve with traditional methods.
%
These two aspects render minimum-time flight in cluttered environments very challenging as any slight disturbance or model mismatch could lead to a catastrophic crash.
To this end, a planning and control method that tackles this task has to be robust against disturbances and needs to be able to adapt the trajectory online.
 

Previous work in the field of trajectory planning and control for autonomous quadrotors has solved only a subset of the problems imposed by minimum-time flight in cluttered environments.  
%
Existing methods either do not consider obstacles in time-optimal planning~\cite{foehn2020cpc, song21RLdroneRacing}, or cannot exploit the full actuation of the platform due to a simplification of the quadrotor dynamics~\cite{liu2018search}.
%
%
Other methods do not support multi-waypoint scenarios in combination with a time-optimal objective~\cite{Webb13_KinodynamicRRTstar}, or rely on polynomial trajectory representations~\cite{richter2016polynomial}, that cannot represent time-optimal maneuvers due to their inherent smoothness.
A recently proposed sampling-based method~\cite{penicka22RALsbmintimequad} can plan minimum-time trajectories in cluttered environments; however, it is decoupled from the model predictive controller (MPC)~\cite{nguyen2021mpcsurvey} used to track the planned trajectory.
This makes flying such a trajectory vulnerable to disturbances or model mismatches as the MPC does not account for the obstacles. 



In light of recent successes in deep reinforcement learning~(RL)~\cite{mnih2015human, song21RLdroneRacing, Fuchs21RLgrandTurismo, lee2020learning, hwangbo2017control}, we propose to address the problem using deep RL.
RL has the advantage of automatically optimizing a parametric controller via trial and error and the ability to handle highly nonlinear dynamical systems and nonconvex objectives that are otherwise intractable to solve by conventional robotics methods. 
However, successful applications of RL have been largely limited to video games~\cite{mnih2015human, Fuchs21RLgrandTurismo}, ground robots~\cite{lee2020learning}, or simple hovering of a quadrotor~\cite{hwangbo2017control}.  
Applying RL to our problem remains a significant challenge due to the high sample complexity and nonconvexity of the task. 

This work contributes a novel learning algorithm for minimum-time flight in cluttered environments. 
The key is to combine classical path planning with model-free deep reinforcement learning to optimize a neural network policy. 
The resulting policy directly outputs optimal control commands from high-dimensional observations.
We show that the neural network policy outperforms state-of-the-art methods in terms of flight time in all tested scenarios that feature complex geometries. 
%
Our results indicate that the learned policy has obtained an implicit knowledge about the risk of navigating in close proximity of obstacles when being exposed to disturbances.
%
This ability leads to more robust control performance and higher success rates when dealing with model mismatches.
%
Our proposed approach is validated in real-world flights at speeds beyond~\SI{42}{\kilo\meter\per\hour} and accelerations up to 3.6g.

\section{Related Work}\label{sec:related}
State-of-the-art methods for agile quadrotor flight mainly use the conventional approach to decouple trajectory planning and control. 
Given a planned trajectory, accurate trajectory tracking by the controller is instrumental for the vehicle to navigate through the environment safely. 
Hence, the final performance and success rate depend highly on both the quality of the planned trajectory as well as the robustness of the controller. 
One of the most popular paradigms to quadrotor trajectory generation exploits the differential flatness~\cite{IJRR2012mellinger_poly} of the platform using polynomial~\cite{richter2016polynomial,burri2015real-time,han2021fastracing,mueller2015TRO_minjerk} or B-spline~\cite{Zhou2020_guided_gradient_planning,zhou2020raptor,Penin2018RAL_vision_reactive_planning} representations.
However, those representations are suboptimal for minimum-time flight, since they are inherently smooth and cannot represent the rapid state or input changes at a reasonable order, and only reach the input limits for infinitesimal short durations~\cite{foehn2020cpc}.
%
%



Search-based planning methods~\cite{Liu_search_based_LQMTC,liu18_searchbased} use discrete-time and discrete-state representations and convert the trajectory planning to a graph search problem.
These methods can optimize time up to discretization, however, they suffer from the curse of dimensionality and utilize simplified point-mass models instead of the full quadrotor dynamics.
%
Furthermore, the existing search-based methods support planning only between two states.
%
Algorithms like RRT*~\cite{Webb13_KinodynamicRRTstar} can be used for a linearized quadrotor model around hover conditions, but the linearization around the hover operating point prohibits planning minimum-time trajectories.

More advanced trajectory planning methods frame the task as a constrained optimization problem and solve it via nonlinear programming.  
As such, trajectory optimization can be used for offline trajectory planning~\cite{foehn2020cpc} or online tracking of a fixed reference path in a receding horizon fashion~\cite{romero2021model}. 
Optimization-based methods have the advantage of being able to incorporate nonlinear dynamics and constraints into the optimization framework.
However, those methods either require long computation times (in the order of hours~\cite{foehn2020cpc}) or rely on a series of approximations for the solver, which in turn results in sub-optimal performance. 
Neither of~\cite{foehn2020cpc, romero2021model} can solve the problem for environments that contain obstacles.
In contrast, the sampling-based method~\cite{penicka22RALsbmintimequad} can find minimum-time trajectories also for cluttered environments.
However, also this method plans a trajectory offline and therefore relies on a controller for tracking. 

Modern control frameworks for trajectory tracking include nonlinear model predictive control~(MPC) and differential flatness control~\cite{sun2021comparative}. 
However, most approaches struggle to handle disturbances during high-speed flight such as aerodynamic drag, thrust mismatches, and system delays. 
When the platform is at its actuation limit, the slightest deviation from the pre-planned trajectory may result in a suboptimal flight path,
and even catastrophic crashes due to the presence of obstacles. 
There exist several MPC approaches that can handle obstacles adaptively online. 
However, they either consider simplified spherical obstacles~\cite{Lindqvist20NMPCobstacles}, or a limited number of obstacles~\cite{small2019NPMCobstalces, Garimella17NMPCobstacles} at a control frequency of only 10-20~\SI{}{\hertz}.
%

Learning-based methods address the aforementioned issues by learning an end-to-end policy that predicts control commands directly from high-dimensional observations.
For example, imitation learning~(IL) methods~\cite{Loquercio21FlightWild, kaufmann2020RSS} train neural network policies that can achieve agile
flight in the wild using only onboard sensing and computing. 
%
IL is data-efficient, but not scalable since it requires designing an expert system for data collection. 
Recent works have demonstrated the usage of reinforcement learning to achieve superhuman performance in car racing~\cite{Fuchs21RLgrandTurismo}, near-time-optimal flight in drone racing~\cite{song21RLdroneRacing}, and high-speed trajectory tracking using a learned policy~\cite{kaufmann2022benchmark}. 
Inspired by~\cite{penicka22RALsbmintimequad, song21RLdroneRacing}, this work combines the topological path planning approach with deep RL to achieve minimum-time flight in complex cluttered environments. 

\section{Problem Statement\label{sec:problem}}


\subsection{Quadrotor Dynamics}

The quadrotor is modeled with state $\bm{x}=\begin{bmatrix} \bm{p},\bm{q},\bm{v},\bm{\omega},\bm{\Omega} \end{bmatrix}^{T}$ which consists of position $\bm{p} \in \Rthree$, velocity $\bm{v} \in \Rthree$, unit quaternion rotation $\bm{q} \in \SOthree$, body rates $\bm{\omega} \in \Rthree$, and  the rotors' rotational speed $\bm{\Omega}$. 
The dynamics equations are
\begin{align}
\label{eq:quat_dyn}
  \begin{aligned}
    \bm{\dot{p}} &= \bm{v} \vphantom{\frac{1}{2}} \\
    \bm{\dot{v}} &= \frac{R(\bm{q})(\bm{f}_{T}+\bm{f}_{D})}{m} + \bm{g}
  \end{aligned}
  &&
  \begin{aligned}
    \bm{\dot{q}} &= \frac{1}{2} \bm{q} \odot \begin{bmatrix} 0 \\ \bm{\omega} \end{bmatrix} \\
    \bm{\dot{\omega}} &= \bm{J}^{-1} (\bm{\tau} - \bm{\omega} \times \bm{J} \bm{\omega}) \vphantom{\frac{1}{2}}
  \end{aligned}
\end{align}
where $\odot$ denotes the quaternion multiplication, $R(\bm{q})$ is the quaternion rotation, $m$ is the mass, $\bm{J}$ is diagonal inertia matrix, and $\bm{g}$ denotes Earth's gravity.
The speeds of the propellers $\bm{\Omega}$ are modeled as a first-order system, ${\bm{\dot{\Omega}} = \frac{1}{k_{mot}}(\bm{\Omega}_{c}-\bm{\Omega})}$ with $\bm{\Omega}_{c}$ being the commanded speed and $k_{mot}$ the time constant. 

The collective thrust $\bm{f}_{T}$ and torque $\bm{\tau}_{b}$ are calculated as:
\begin{equation}
\label{eq:tau_thrust}
\bm{f}_{T}=\begin{bmatrix} 0 \\ 0 \\ \sum f_i \end{bmatrix} \text{, }
\bm{\tau} = 
\begin{bmatrix} 
 l/\sqrt{2}(f_{1}-f_{2}-f_{3}+f_{4})   \\
 l/\sqrt{2}(-f_{1}-f_{2}+f_{3}+f_{4})   \\
 \kappa (f_{1}-f_{2}+f_{3}-f_{4})
\end{bmatrix} 
\text{,}
\end{equation}
where $\kappa$ is the torque constant and $l$ is the arm length.
Here, individual motor thrusts $f_{i}$ are functions of the motor speeds using the thrust coefficient $c_{f}$ as in~\eqref{eq:thrusts},
\begin{equation}
\label{eq:thrusts}
f_{i}(\Omega) = \begin{bmatrix} c_f \cdot \Omega^{2} \end{bmatrix} \text{. }
\end{equation}

The drag force $\bm{f}_{D}$ is modeled as a linear function of velocity in body frame $\bm{v}_{\mathcal{B}}$~\cite{faessler2017differential} with drag coefficients $(k_{vx},k_{vy},k_{vz})$: 
\begin{equation}
\label{eq:drag}
\bm{f}_{D} = - \begin{bmatrix}k_{vx} v_{\mathcal{B},x} & k_{vy} v_{\mathcal{B},y} & k_{vz} v_{\mathcal{B},z}\end{bmatrix}^{T} \text{.}
\end{equation}

The motors have a limited thrust range $[f_{min},f_{max}]$: 
\begin{align}
  f_{min} \leq &f_{i} \leq f_{max} \text{, for } i \in \{1,\ldots,4\} \; . \label{eq:motor_constraints} 
\end{align}

\subsection{Minimum-time Planning Problem}

The minimum-time planning problem is defined using the classical notion of configuration space $\cspace$~\cite{lavalle2006planning}.
We assume an environment $\world = \Rthree$, which contains obstacles $\obstacles = \{\obstacles_{1},\dots,\obstacles_{m}\} \subset \world$.
The quadrotor with state $\bm{x}$ and geometry $\robot(\bm{x}) \subset \world$ has to find a collision-free trajectory in $\cspace$.
To this end, it can move in free space $\Cfree = \cspace \setminus \Cobs$, where $\Cobs = \{ \bm{x} \in \cspace | \delta(\robot(\bm{x}),\obstacles) \leq d_c \} \subseteq \cspace$ is a set of configurations (quadrotor states) where the robot is in collision, i.e., having shortest distance $\delta(\cdot,\cdot)$ to any obstacle below a given threshold $d_c$.

We formulate the multi-waypoint minimum-time planning problem as an optimization problem~(\ref{objective}) to find trajectories $\tau_{i}$ and their durations $t_i$ for the dynamics \eqref{eq:quat_dyn}-\eqref{eq:motor_constraints}.
\begin{equation} \label{objective}
   \begin{split}
      \minimize_{\tau_0 \ldots \tau_N}~& T=\sum_{i=0}^{N} t_i    \\
      \subjectto \text{ }& \tau_i \in \Cfree \text{ for } i \in \{0,\ldots,N\}\text{,}\\
      & \tau_{0}(0) = \bm{x}_{s} \text{, } \tau_{N}(1) = \bm{x}_{e}\text{,}\\
      & \norm{\tau_{i}(0)_{p} - \bm{p}_{wi}} \leq r_{tol}\text{ for } i \in \{1,\ldots,N\}\text{,}\\
      & \tau_{i-1}(1) = \tau_{i}(0) \text{ for } i \in \{1,\ldots,N\} \text{,}\\
      & \eqref{eq:quat_dyn}\text{, } \eqref{eq:tau_thrust}\text{, } \eqref{eq:thrusts} \text{, } \eqref{eq:drag} \text{, } \eqref{eq:motor_constraints}\text{.} 
   \end{split}
\end{equation}

For the multi-waypoint scenario, the quadrotor has to fly through a given sequence of waypoints $P_w = (\bm{p}_{wi}, i \in [1,\ldots,N])$ while reaching their position $\bm{p}_{wi}$ with a certain proximity $r_{tol}$. 
The whole multi-waypoint trajectory can be described as a continuous sequence of $N$ trajectories $\tau_i: [0,1] \to \Cfree$ for $i \in \{0,\ldots,N\}$. 
The trajectory is assumed to have a given start $\tau_{0}(0) = \bm{x}_{s}$ and end $\tau_{N}(1) = \bm{x}_{e}$.
Furthermore, the initial positions of the trajectories $\tau_{i}(0)_{p}$ have to be in the waypoints' proximity $\norm{\tau_{i}(0)_{p} - \bm{p}_{wi}} \leq r_{tol}$ for $i \in \{1,\ldots,N\}$, and the sequence has to be continuous $\tau_{i-1}(1) = \tau_{i}(0)$ for $i \in \{1,\ldots,N\}$.
The goal of the planning problem is then to minimize the final time $T=\sum_{i=0}^{N} t_i$ of reaching $\bm{x}_{e}$, where $t_i$ denotes the time duration of $\tau_{i}$.


\section{Methodology\label{sec:method}}

\begin{figure*}[!htb]
   \centering
  \setlength\tabcolsep{0.5pt}
  \begin{tabular}{lcr}
  \includegraphics[width=0.33\linewidth]{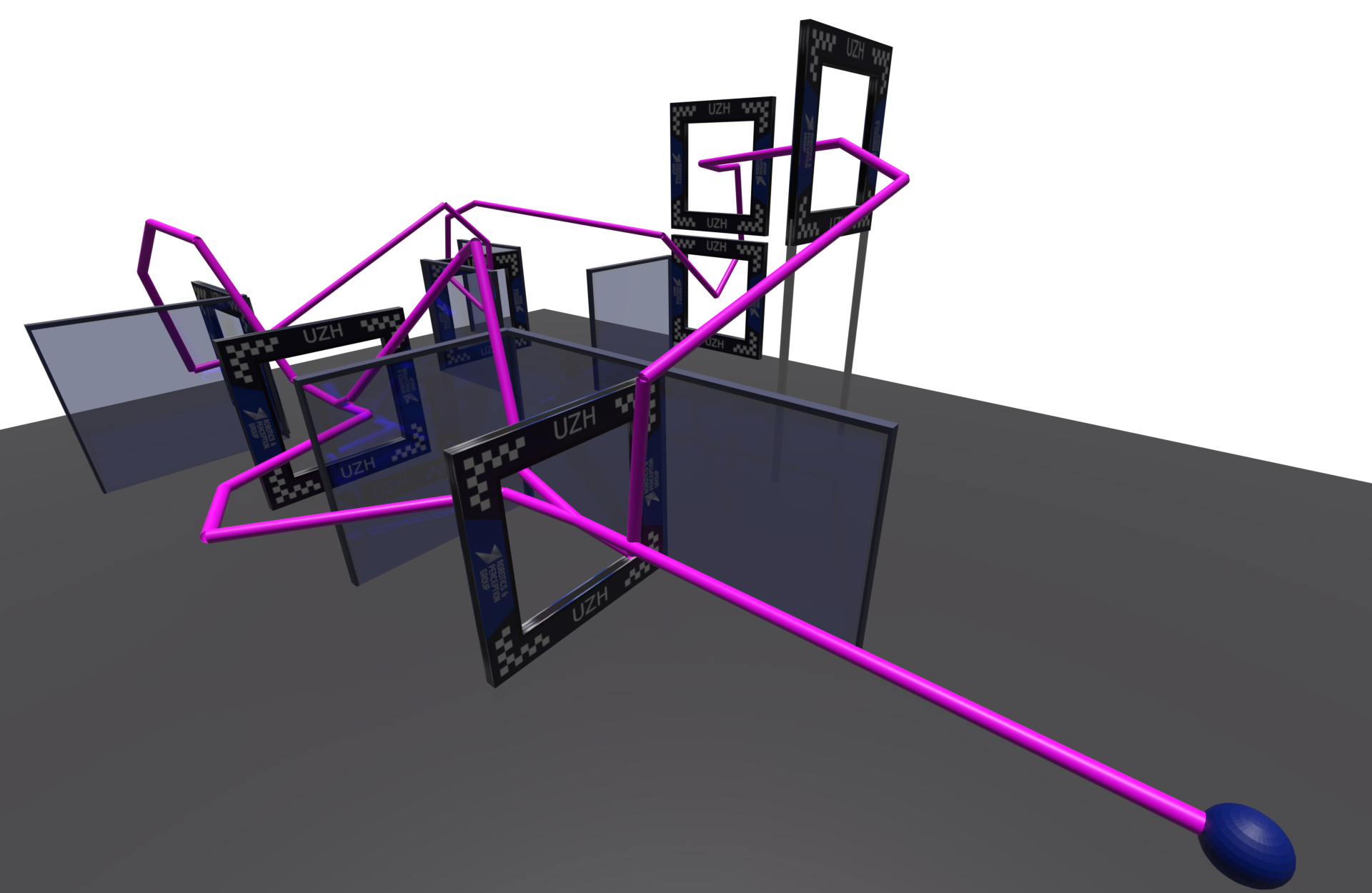}
  &
  \begin{tikzpicture}[      
        every node/.style={anchor=south west,inner sep=0pt},
        x=1mm, y=1mm,
      ]   
     \node at (0,0) {\includegraphics[width=0.333\linewidth]{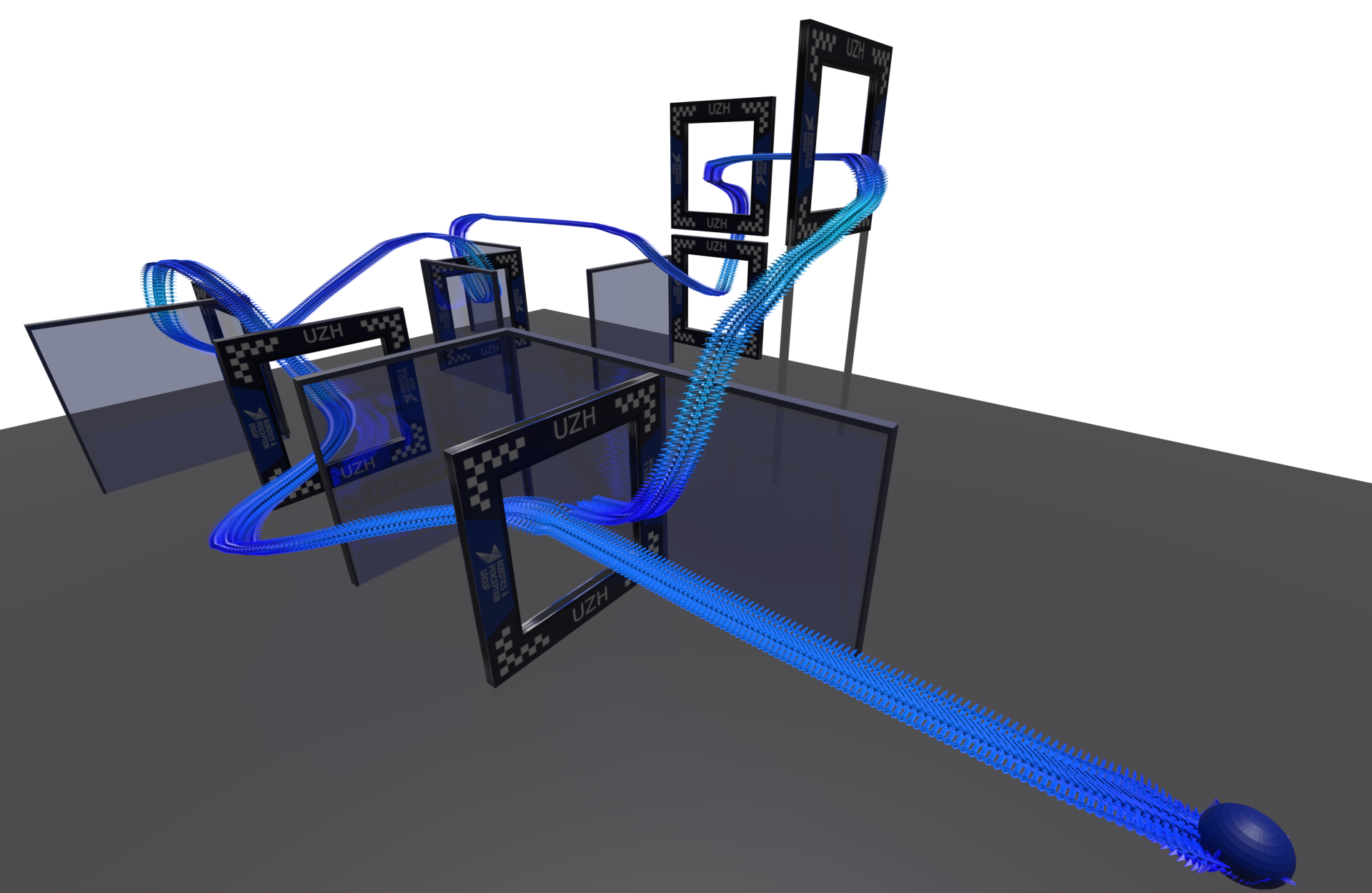}};
     \node at (47,11) {\includegraphics[height=0.15\linewidth]{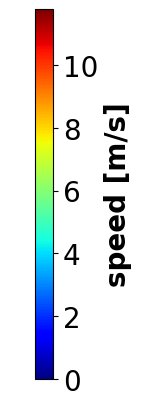}};
  \end{tikzpicture}
  &
  \noindent
  \begin{tikzpicture}[      
        every node/.style={anchor=south west,inner sep=0pt},
        x=1mm, y=1mm,
      ]   
     \node at (0,0) {\includegraphics[width=0.33\linewidth]{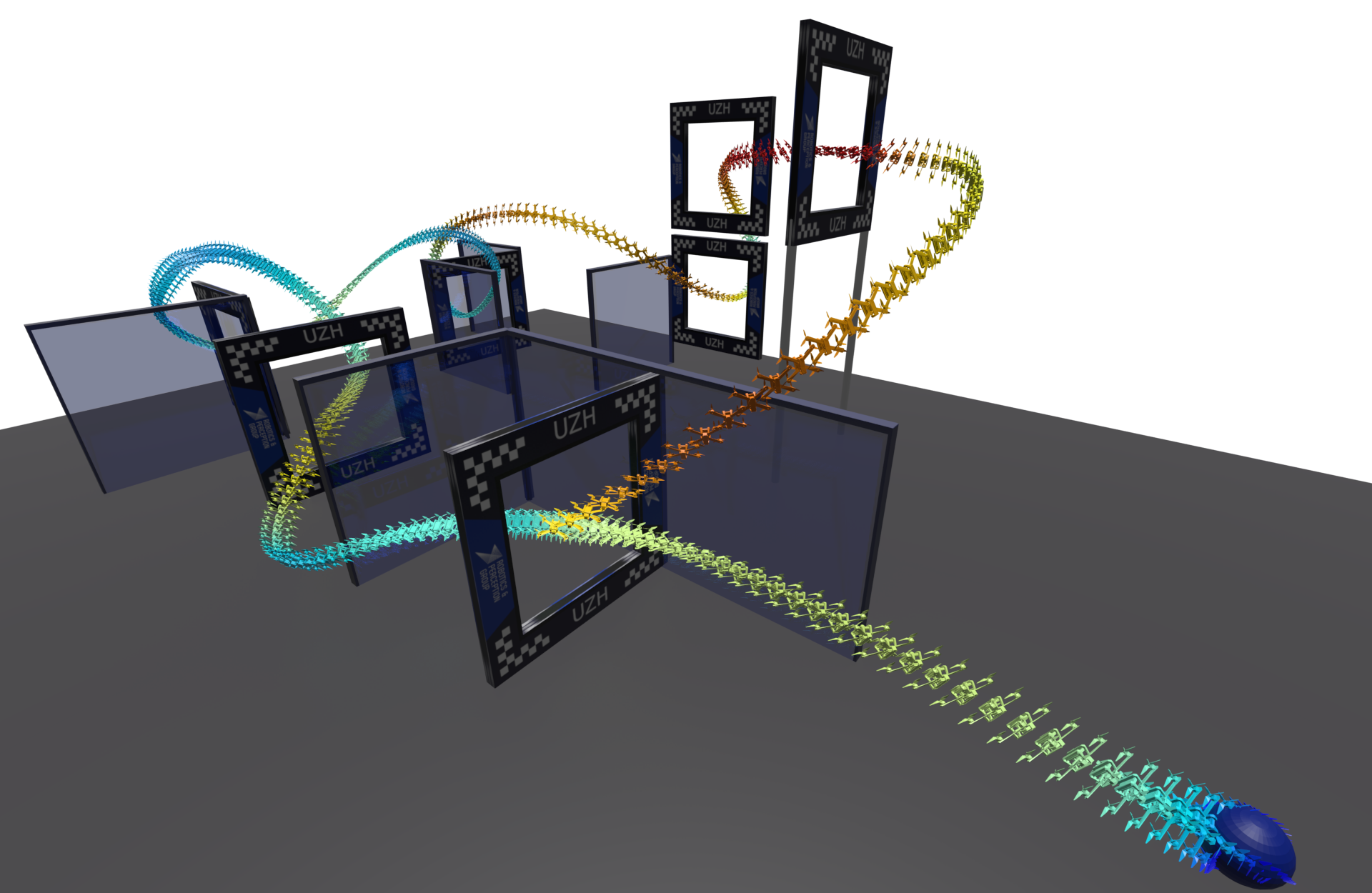}};
     \node at (47,11) {\includegraphics[height=0.15\linewidth]{fig/image_colorbar.png}};
  \end{tikzpicture}
  \\[-1.5em]
 \multicolumn{1}{l}{\color{white} ~(a) Topological guiding paths} 
 &
 \multicolumn{1}{l}{\color{white} ~(b) Slow speed trajectory} 
 & 
 \multicolumn{1}{l}{\color{white} ~(c) Final minimum-time trajectory}
   \end{tabular}
   \vspace{-0.5em}
   \caption{
   Three main steps of our method (shown in Slalom environment) include: (a) finding topological guiding paths between the waypoints (gates), (b) learning slow policy that flies through all the waypoints, and (c) learning minimum-time policy.
      \label{fig:methodstages} 
    \vspace{-1.7em}
    }
   
\end{figure*}

\disable{
The proposed method follows three main steps as shown in Figure~\ref{fig:methodstages}.
First, collision-free topological guiding paths between the specified waypoints are found using a version of PRM~\cite{penicka22RALsbmintimequad} to capture the connectivity of the free space between the waypoints.
The guiding paths are used to both focus the reinforcement learning in their close vicinity, and to calculate reward based on progress along the paths.
In the second step, we learn slow trajectories that are only allowed to fly close to the guiding paths, by penalizing high-speed motions and distance from the  guiding path, until the quadrotor can fly through all waypoints using the learned policy.
This allows to find an initial feasible trajectory for given dynamics \eqref{eq:quat_dyn}-\eqref{eq:motor_constraints} easier, as the narrow passages would become smaller at high speeds, and also enables to store feasible states along the topological paths that are further used for random agent initialization during learning.
Finally, the speed limits and the limited distance from the guiding path are removed to learn high-speed minimum-time policy. 
}

The key ingredients of our approach to minimum-time flight in cluttered environments are three-fold: 
1)~generation of a topological guiding path using a probabilistic roadmap~\cite{TRA96KavrakiPRM}, 
2)~a~novel task formulation that combines progress maximization along the guiding path with obstacle avoidance, 
and 3)~a~curriculum training strategy to train a neural network policy using deep reinforcement learning.

\subsection{Topological Path Planning\label{subsec:topo_paths}}
We first find the topological paths that connect individual waypoints to guide the subsequent learning process.
The topological guiding paths are found using a variant of the Probabilistic Roadmap~\cite{TRA96KavrakiPRM} described in~\cite{penicka22RALsbmintimequad}. 
The algorithm searches for multiple distinct paths with different homotopy classes, e.g., going around obstacles from different sides.
It uses random sampling in ellipsoids between individual waypoints and keeps enlarging each ellipsoid and number of samples until at least one path between waypoints is found.

The created roadmaps between waypoints are then searched for the shortest path using Dijkstra's algorithm.
Samples within the shortest path with the smallest distance to obstacles are then removed from the roadmap and the shortest path search is repeated to obtain multiple distinct paths. 
Such distinct paths are then shortened, and the paths within the same homotopy class are removed.
The resulting topological paths then represent the connectivity of the $\Cfree$ between waypoints (as shown in Fig.~\ref{fig:methodstages}(a)).
For more details about the topological path search, we refer to~\cite{penicka22RALsbmintimequad}.






\subsection{Reinforcement Learning for Minimum-time Flight}
In the following, we present the policy architecture, reward formulation, and strategy employed in our approach to train a policy for high-speed flight in cluttered environments. 

\subsubsection{Policy Architecture}
%
The neural network policy uses an observation space that consists of three main parts: the quadrotor state, the next waypoint position, and the relative position of the farthest collision-free position on the guiding path. 
Specifically, we denote the observation vector as ${\bm{o}=[\bm{p}(t),R(\bm{q}(t)),\bm{v}(t),\mathcal{W},\bm{\gamma}(t)]}$, where $\bm{p}(t)$, $R(\bm{q}(t))$, $\bm{v}(t)$ are the quadrotor's position, rotation matrix and velocity, respectively.
Matrix $\mathcal{W} \in \mathcal{R}^{4,3}$ contains four positions corresponding to the bounding box of the currently targeted waypoint with edge size of $2r_{tol}$.
The bounding box is used to convey information not only about the waypoint position, but also about the size and orientation of the target gate (see Fig.~\ref{fig:illustration}).
Finally, the $\bm{\gamma}(t)$ is the farthest point on the guiding path connectable using a collision-free line segment from the current position $\bm{p}(t)$ (see Fig.~\ref{fig:progress}).
The $\bm{\gamma}(t)$ is used for indicating the flight direction of the quadrotor to maximize the progress.
Throughout the development of the method, we found by ablating the observation components that all the components are necessary to learn minimum-time flight in all tested scenarios.

\begin{figure}[!t]
\centering
  \includegraphics[width=0.9\linewidth]{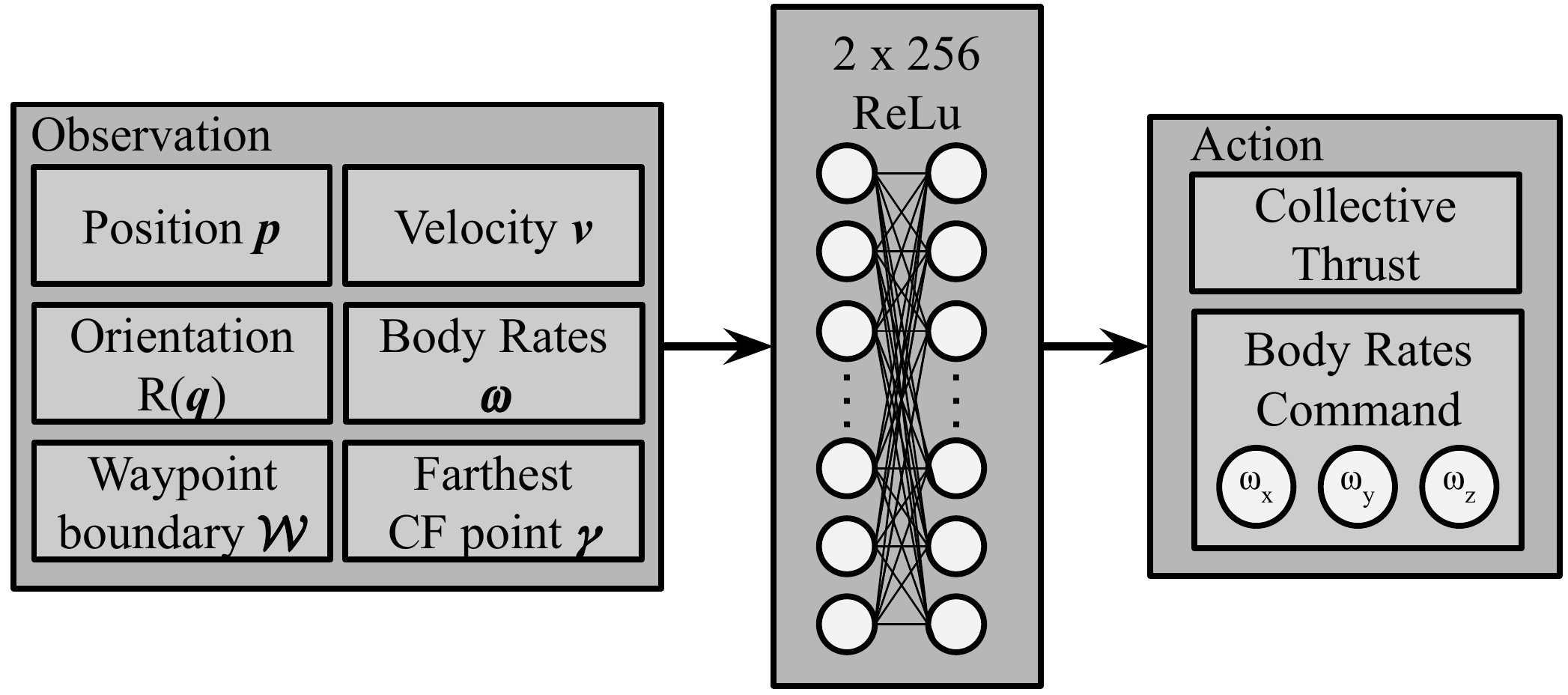}
  \vspace{-0.26cm}
  \caption{
  Illustration of the observation and action produced by the neural network policy.
  \label{fig:rl_illustration}
  \vspace{-1.4em}}
\end{figure}

The action produced by the policy $\bm{a}(t)=[f_T,\bm{w}]$ is the collective thrust $f_T$ in body-z axis and the commanded body rates $\bm{w}$.
This action modality has been identified in~\cite{kaufmann2022benchmark} as the best performing for learning-based control policies.
A low-level controller then tracks the desired collective thrust and body rates to produce the speed command for each rotor.
Figure~\ref{fig:rl_illustration} illustrates the observation and action spaces, as well as the neural network architecture, which is a 2-layer multilayer perception~(MLP).  

\subsubsection{Path Progress Maximization and Obstacle Avoidance}
%
The objective of the studied problem is to minimize the time of reaching the last waypoint, which however, represents a very sparse signal that is difficult to optimize.
This is why prior works opted for a dense proxy reward using the projected progress along the center line of a race track in car racing~\cite{Fuchs21RLgrandTurismo}, or progress along the straight line segments between gates~\cite{song21RLdroneRacing} for the task of drone racing.
We further extend this progress-based reward for the task of minimum-time planning in cluttered environments by calculating the progress along the topological guiding path.

\begin{figure}[!t]
\centering
   \includegraphics[width=1.0\linewidth,trim={0 1.5cm 0 0.1cm},clip]{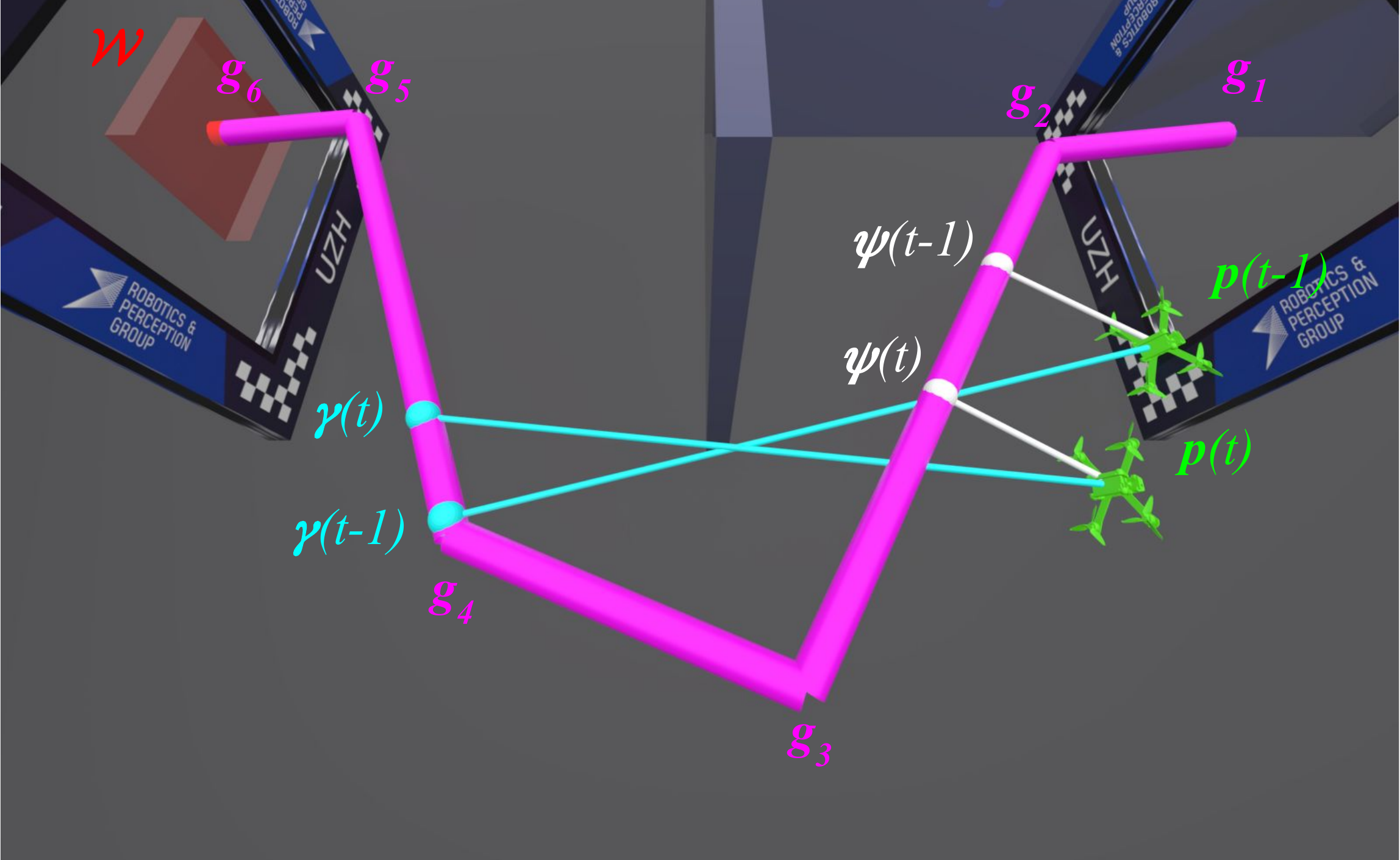}
   \vspace{-0.6cm}
   \caption{
   Illustration of the collision-free topological guiding path between waypoints. 
   The nearest point on the guiding path $\bm{\psi}$ from quadrotor position $\bm{p}$ is used to calculate progress reward.
   The farthest collision-free point $\bm{\gamma}$ and waypoint bounding box  $\mathcal{W}$ are used as a part of observation.
      \label{fig:progress}
      \vspace{-2.0em}
      }
\end{figure}

Figure~\ref{fig:progress} illustrates how the progress is computed between two consecutive states with positions $\bm{p}(t-1)$ and $\bm{p}(t)$.
We assume the guiding topological path between start waypoint and end waypoint consists of a sequence of $n$ points $(\bm{g}_1,\ldots,\bm{g}_{n})$ that form a sequence of line segments $(l_1,\ldots,l_{n-1})$.
To calculate the progress at position $\bm{p}$, we need to find the closest point $\psi(\bm{p})$ on the guiding path and its line segment index $l(\bm{p})$ as: 
\begin{equation}
 \begin{split}
l(\bm{p}), \bm{\psi}(\bm{p}) &= \argmin_{l(\bm{p}),\psi(\bm{p})} \norm{\bm{p}-\bm{\psi}(\bm{p})}\\
\subjectto \text{ }&  \bm{\psi}(\bm{p}) = \bm{g}_{l(\bm{p})} + t (\bm{g}_{l(\bm{p})+1} - \bm{g}_{l(\bm{p})}) \text{,} \\
 & t = \frac{(\bm{p}-\bm{g}_{l(\bm{p})}) \cdot (\bm{g}_{l(\bm{p})+1}-\bm{g}_{l(\bm{p})}) }{ \norm{\bm{g}_{l(\bm{p})+1}-\bm{g}_{l(\bm{p})}}^{2}}\text{,} \\
 &l(\bm{p}) \in \{1,\ldots,n-1\} \text{ , } t \in [0,1] \text{.} 
 \end{split}
\end{equation}

The reached distance $s(\bm{p})$ along the guiding path is then calculated using~\eqref{eq:projection} as the length of the topological path until the closest point $\bm{\psi}(\bm{p})$.
The progress reward $r_{p}(t)$ at time $t$ is then computed using equation~\eqref{eq:progress} as a difference in reached distance between the current and previous time step.
\begin{align}
    \label{eq:projection}
    s(\bm{p}) &= \sum_{i=1}^{l(\bm{p})-1} \norm{\bm{g}_{i+1} - \bm{g}_{i}} + \norm{\bm{\psi}(\bm{p})-\bm{g}_{l(\bm{p})}} \\
    \label{eq:progress}
    r_{p}(t) &= s(\bm{p}(t)) - s(\bm{p}(t-1)) 
\end{align}

The total reward $r(t)$ at time $t$ then equals
\begin{equation}
    \label{eq:reward}
    r(t) = k_{p}r_{p}(t) + k_{s}s(\bm{p}(t)) + k_{wp}r_{wp} + r_{T} - k_{\omega}\norm{\bm{\omega}}\text{,}
\end{equation}
where $k_{p}$, $k_{\omega}$, and $k_{wp}$ are hyperparameters that define the contribution of each reward component. 
The reached distance $s(\bm{p}(t))$ with parameter $k_{s}$ is used as part of the reward mainly to counteract the fact that the progress along the line segments can have many singularities.
Such singularities emerge due to the sharp corners of the guiding path (see Fig.~\ref{fig:progress}) and minimum-distance projection, which can block learning a policy that flies through all waypoints when negative progress occurs in the singularities.

The total reward also values passing of a waypoint $k_{wp}r_{wp}$ and discourage high body rates $k_{\omega}\norm{\bm{\omega}}$.
When a new waypoint is passed within distance $d_{w} \leq r_{tol}$ a positive reward of $r_{wp}=e^{-d_{wp}/r_{tol}}$ is added to prioritize passing close to the waypoint center and thus to increase robustness during real flight with disturbances.
The terminal reward $r_{T}=-10$ is only added when the quadrotor collides with an obstacle. 
During the development of the method, we tested several reward components and their ablation, rendering only the presented variant able to learn policies for all tested scenarios.

\subsubsection{Training Strategy}
Naive optimization of the reward formulation specified in~\eqref{eq:reward} results in suboptimal performance due to local minima and a higher probability of collisions in high-speed flight. 
This is caused by the decoupled nature of the topological path planning and the reinforcement learning of minimum-time flight for the dynamic quadrotor model.
Such decoupling makes the learning of high-speed flight along the topological paths a challenging problem.

We overcome this limitation by employing a curriculum strategy, where the racing policy is trained in two stages. 
In the first stage, a slow policy (see Fig.~\ref{fig:methodstages}(b)) is trained to fly closely along the guiding path while the minimal $v_{min}$ and maximal $v_{max}$ speeds are limited through a scaled reward.
This helps to find a policy that flies through the waypoints without collision as the guiding paths are known to be collision free and the high velocities would decrease the relative size of narrow passages.
Specifically, in the initial slow flight training stage, the parameters of progress reward $k_{p}$ and reached distance reward $k_{s}$ are scaled down by a factor of $s$ computed as:
\begin{align}
&s = s_{v_{max}} s_{v_{min}} s_{gd} \text{,}\\
&s_{v_{max}} = 
\begin{cases}
    10^{v_{max}-\norm{\bm{v}}} & \text{if } \norm{\bm{v}} > v_{max}\text{,}\\
    1 & \text{ otherwise,}
\end{cases}\\
&s_{v_{min}} = \begin{cases}
    10^{\norm{\bm{v}}-v_{min}} & \text{if } \norm{\bm{v}} < v_{min}\text{,}\\
    1 & \text{ otherwise,}
\end{cases}
\\
&s_{gd} = \begin{cases}
    e^{-\norm{\bm{p}-\bm{\psi}(\bm{p})} + d_{max}} & \text{if }  \norm{\bm{p}-\bm{\psi}(\bm{p})} > d_{max}\text{,}\\
    1 & \text{ otherwise.}
\end{cases}
\end{align}

This adapted reward formulation forces the policy to find only slow trajectories with speed $v_{min} < \norm{\bm{v}} < v_{max}$ and distance from the guiding paths $\norm{\bm{p}-\bm{\psi}(\bm{p})} < d_{max}$, which in turn helps finding a trajectory around an already known collision-free guiding path.

The limited maximal speed in the slow flight training  stage is also used to calculate the initial value of the parameter $k_{s}$ as ${k_{s} = 2(v_{max} \cdot d_{t})/\sum_{i=1}^{n-1} \norm{\bm{g}_{i+1} - \bm{g}_{i}}}$, where $d_{t}$ is the simulation time step.
This limits the collected reward from reached distance to be approximately the same as the progress reward in the slow flight learning phase.
While during the later minimum-time learning phase, the progress reward is, in comparison, significantly more prominent. 

After the trained slow flight policy is able to navigate through all waypoints, the speed limits and the constrained distance to the guiding path are removed to enable training a minimum-time policy as shown in Fig.~\ref{fig:methodstages}(c).

\subsubsection{Training Details}
The policy is trained using Proximal Policy Optimization (PPO)~\cite{schulman2017proximal}, which has demonstrated good performance in benchmarks for continuous control tasks.
We utilize 100 parallel agents to train the policy entirely in simulation, which increases the speed of collecting data and diversifies the experienced states and observations among agents.
The simulation uses the dynamics \ref{eq:quat_dyn}-\ref{eq:motor_constraints} and forward integrates them using a 4th order Runge-Kutta scheme.
Additionally, the linear drag coefficients $(k_{vx},k_{vy},k_{vz})$ are randomized with normal distributions $N(0,k_{vx}) ,N(0,k_{vy}) ,N(0,k_{vz})$ for each agent after restart, to make the policy robust against unknown and possibly random aerodynamic effects.

Initialization of the agents is randomized among all specified waypoints and guiding paths to encourage diversity of experienced states.
Each agent keeps a vector of valid states, i.e., states that are reached by running the policy from the start position without collision.
In the first learning stage, the valid states are required to be within the speed limit $v_{min} < \norm{\bm{v}} < v_{max}$ and close to the guiding path $\norm{\bm{p}-\bm{\psi}(\bm{p})} < d_{max}$.
The guiding paths are discretized into \SI{1}{\meter} parts based on the distance along the path $s(\bm{p})$, and each such part is mapped to one valid state in the vector of agent's valid states.
These valid states are then used to randomly initialize the state after an agent collides with an obstacle or reaches the final waypoint.

All policies are trained using 12 threads on a laptop featuring an Intel Xeon W-10885M CPU 
and a Quadro RTX 4000 Mobile GPU.

\section{Results\label{sec:results}}

\begin{figure*}[!htb]
   \centering
   \setlength\tabcolsep{-3.5pt}

  \begin{tabular}{ccc}
  \subcaptionbox{Forest\label{fig:columnts}}{
  \begin{tikzpicture}[      
        every node/.style={anchor=south west,inner sep=0pt},
        x=1mm, y=1mm,
      ]   
     \node at (0,0) {\includegraphics[width=0.33\linewidth]{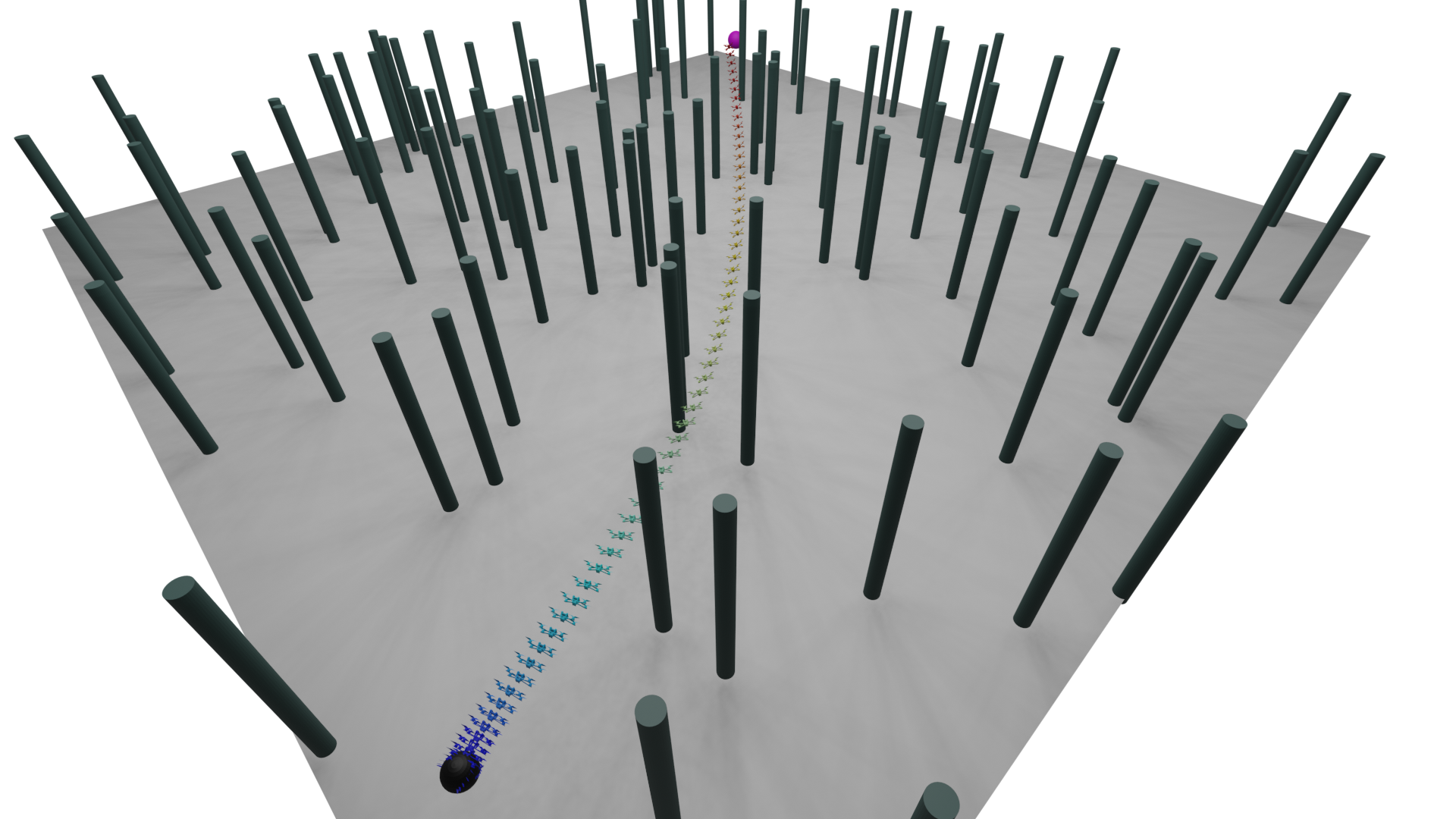}};
     \node at (0,7) {\includegraphics[height=0.15\linewidth]{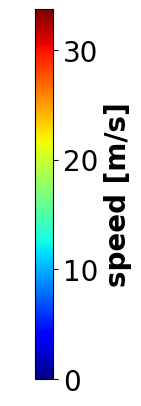}};
  \end{tikzpicture}
  \vspace{-0.5em}
  }   &  
  \subcaptionbox{Office\label{fig:office}}{
  \begin{tikzpicture}[      
        every node/.style={anchor=south west,inner sep=0pt},
        x=1mm, y=1mm,
      ]   
     \node at (0,0) {\includegraphics[width=0.33\linewidth]{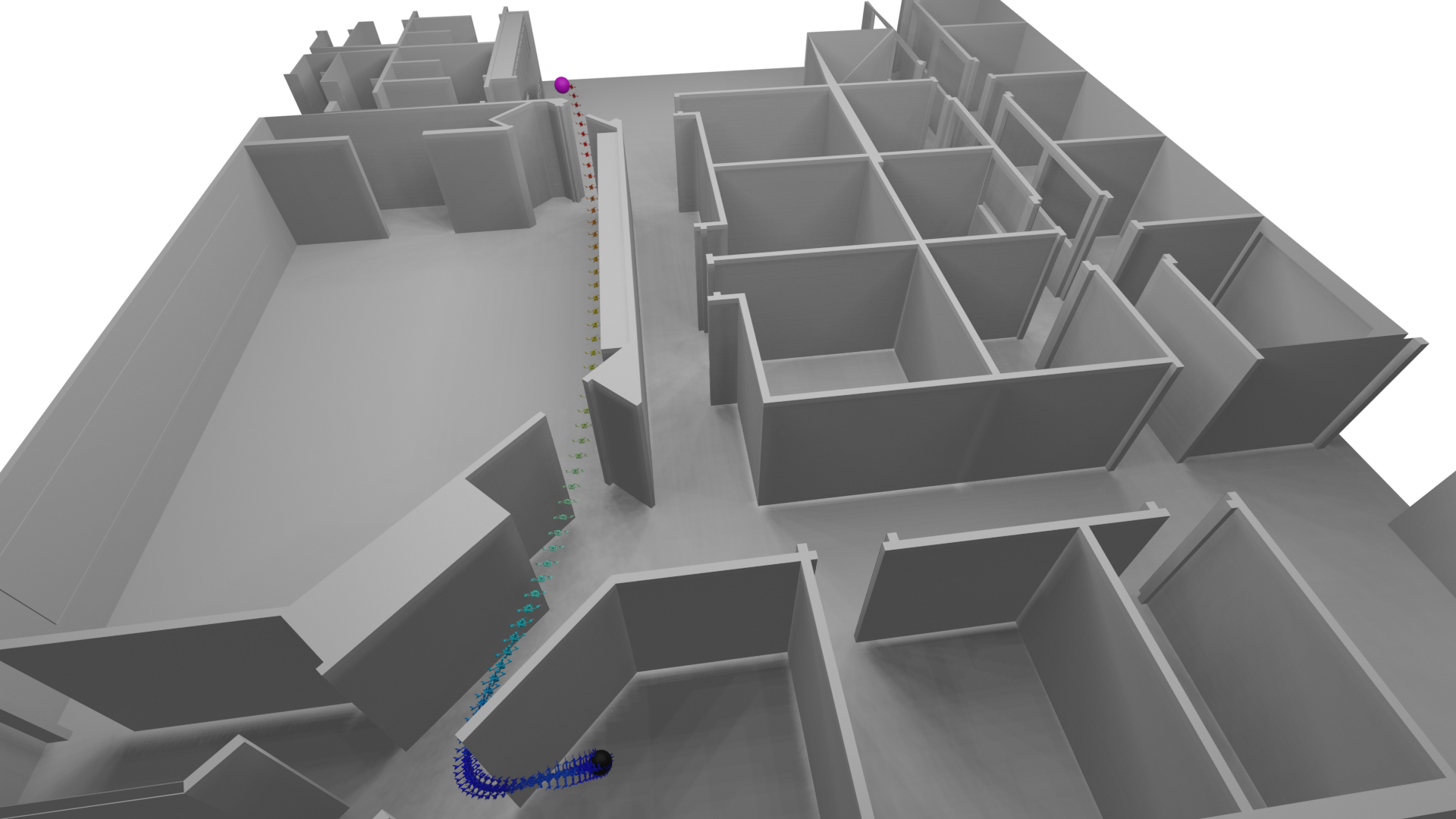}};
     \node at (0,7) {\includegraphics[height=0.15\linewidth]{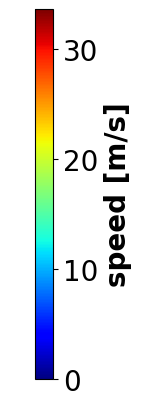}};
  \end{tikzpicture}
  \vspace{-0.5em}
  }   
    &
  \subcaptionbox{Racing / Racing MW\label{fig:racing}}{
  \begin{tikzpicture}[      
        every node/.style={anchor=south west,inner sep=0pt},
        x=1mm, y=1mm,
      ]   
     \node at (0,0) {\includegraphics[width=0.33\linewidth]{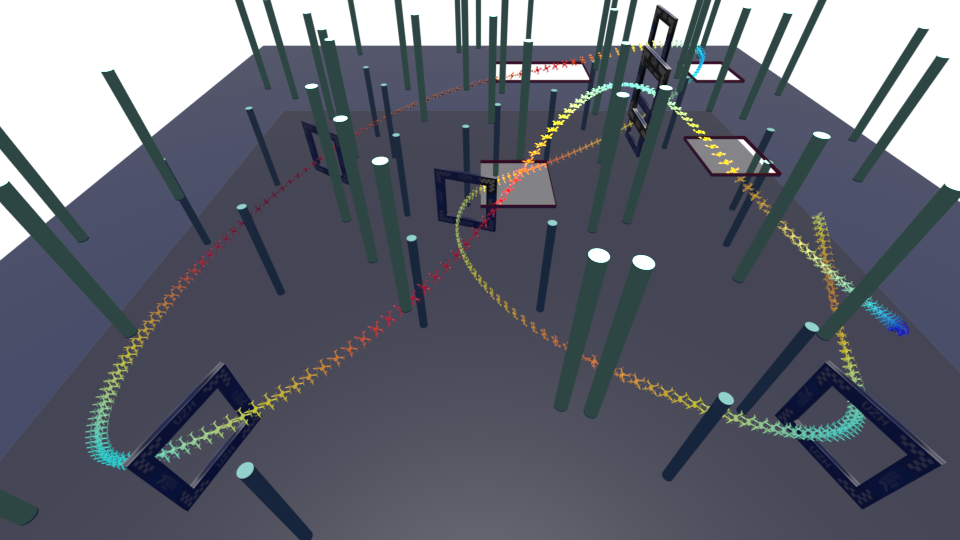}};
     \node at (0,7) {\includegraphics[height=0.15\linewidth]{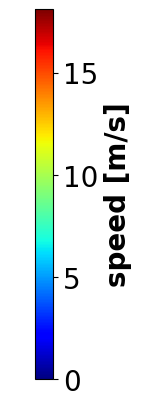}};
  \end{tikzpicture}
  \vspace{-0.5em}
  }   
   \end{tabular}
    \vspace{-0.9em}
   \caption{
   Example trajectories found by the proposed method in the selected environments used for evaluation.
      \label{fig:environments} 
    \vspace{-1.7em}
    }
\end{figure*}


The proposed method has been evaluated concerning the following performance criteria: (i) lap time performance of the planned trajectory in a simple simulation scenario without model mismatch, (ii) success rates of navigating the vehicle through a cluttered environment using high-fidelity simulation, and (iii) validation of the learned policy in the real world.

We identify a powerful race drone using real world flight data. 
The physical properties of the quadrotor, along with hyperparameters of the proposed and baseline methods, are summarized in Table~\ref{tab:config}.
The hyperparameters of the reward components were tuned such that the collected rewards from the individual components follow a particular priority, i.e. $-r_{T} \gg k_{p}r_{p}(t) \approx k_{wp}r_{wp} \gg k_{s}s(\bm{p}(t)) \approx - k_{\omega}\norm{\bm{\omega}}$.
The evaluation of the method is done in four environments.
The Slalom environment is shown in Figure~\ref{fig:methodstages}.
The Forest, Office, and Racing environments, including the Multi-Waypoint~(MW) Racing environment, are illustrated in Figure~\ref{fig:environments}.
The environments are represented as an Euclidean Signed Distance Field~\cite{park2019deepsdf} with precision of \SI{0.05}{\meter}. 
We use Flightmare~\cite{song2020flightmare} for the simulation and the Stable Baselines~\cite{stable-baselines3} for policy training.

\begin{table}[!htb] 
   \centering 
   \footnotesize 
   \renewcommand{\tabcolsep}{4.4pt} 
   \renewcommand{\arraystretch}{0.8} 
   \caption{Parameters of the quadrotor and the algorithms\label{tab:config}. }  
   \vspace{-1em} 
   \begin{tabular}{c|cc|cc}
     \toprule 
  & Variable & Value & Variable & Value \\
       \midrule 
   \parbox[t]{2mm}{\multirow{6}{*}{\rotatebox[origin=c]{90}{Quadrotor}}} & $m$ [\SI{}{\kg}] & 0.85 & $l$ [\SI{}{\meter}] & 0.15 \\
   & $f_{min}$ [\SI{}{\newton}] & 0 & $f_{max}$ [\SI{}{\newton}] & 7 \\
   & $\text{diag}(J)$ [\SI{}{\gram\metre\squared}] & $[1,1,1.7]$ & $\kappa$ [-] & 0.05 \\
   & $w_{max}$ [\SI{}{\radian\per\second}] & 15 & $c_{f}$ & \num{1.563e-6} \\
   & $k_{vx}$ [\SI{}{\newton\second\per\meter}] & 0.26 & $k_{vy}$ [\SI{}{\newton\second\per\meter}]& 0.28 \\
   & $k_{vz}$ [\SI{}{\newton\second\per\meter}]& 0.42 & & \\
   \midrule
   \cite{richter2016polynomial} & $k_{T}$ & $105000$ & $N_{poly}$ & 10 \\[-0.2em] 
   \midrule
   \cite{Liu_search_based_LQMTC} & $a_{max}$ [\SI{}{\meter\per\second\squared}] & 31.4 & & \\
    \midrule
   \parbox[t]{2mm}{\multirow{4}{*}{\rotatebox[origin=c]{90}{RL}}} 
   & $k_{p}$[-] & 5.0 & $k_{\omega}$[-]& 0.01\\
   & $k_{wp}$[-] & 5.0 & $d_{t}$ [\SI{}{\second}] & 0.02  \\ 
   & $v_{max}$ [\SI{}{\meter\per\second}] & 2 & $v_{min}$ [\SI{}{\meter\per\second}] & 1 \\
   & $d_{max}$ [\SI{}{\meter}]& 0.3 & $d_{c}$ [\SI{}{\meter}] & 0.15 \\
    \bottomrule 
   \end{tabular}  
\end{table} 

\subsection{Lap Time Performance of Planned Trajectories in Simple Simulation}

In the simple simulation, we analyze the quality, such as duration, of the planned trajectories and compare it with related baseline algorithms.
The first baseline algorithm is the polynomial method~\cite{richter2016polynomial} that jointly minimizes snap and time (using time penalty $k_{T}$) of a trajectory  that is represented by an $N_{poly}$-th order polynomial.
The search-based method~\cite{Liu_search_based_LQMTC} for quadrotor planning uses a discretized state of a point-mass up to acceleration to plan minimum-time trajectories using a graph-search algorithm.
The search-based method uses an acceleration limit of $a_{max}=31.4~\SI{}{\meter\per\second\squared}$, which represents the optimistic, yet infeasible, limit corresponding to achievable acceleration along one axis in horizontal motion. 
The optimization method presented in~\cite{foehn2020cpc} plans truly time-optimal trajectories between given waypoints, however, the original variant had to be extended to allow planning in cluttered environments.
Lastly, the sampling-based method~\cite{penicka22RALsbmintimequad} uses a hierarchical approach of increasing model complexity to plan minimum-time trajectories for the full quadrotor model in cluttered environments. 

To allow for a fair comparison, the motor dynamics $\dot{\Omega}$ and the aerodynamic drag forces $\bm{f}_{d}$ are not considered in the simple simulation.
This corresponds to the baseline methods~\cite{richter2016polynomial,Liu_search_based_LQMTC,foehn2020cpc,penicka22RALsbmintimequad}, which all do not account for both phenomena, except for ~\cite{foehn2020cpc} that can include a simple linear drag model.
The comparison of the planning methods is presented in Table~\ref{tab:single_target}.
The trajectory duration is reported as the best-found duration $T_b$.
Additionally, we show the average duration with standard deviation $T_{a}$ from 30 different runs for the methods that are randomized and have non-zero deviation.
We report the computation times for the baseline methods. 
For the proposed method, we show the inference time (including the time for preprocessing observations) of the trained neural network during evaluation. 
For each scenario, a different policy is learned with training time of approximately 45 minutes with a standard laptop.

\begin{table*}[t] 
   \centering 
   \footnotesize 
   {\renewcommand{\tabcolsep}{4.5pt} 
   \caption{Comparison of baseline algorithms and our learning-based method.\label{tab:single_target}} 
   \vspace{-1em} 
   \begin{tabular}{lcrrrrrcrrrrrrr}
     \toprule 
    \multirow{2}{*}{Environment} & \multirow{2}{*}{\begin{minipage}{5mm}Test\\case\end{minipage}} & \multicolumn{3}{c}{Polynomial~\cite{richter2016polynomial}} &   \multicolumn{2}{c}{Search-based~\cite{Liu_search_based_LQMTC}} &  \multicolumn{2}{c}{CPC~\cite{foehn2020cpc}} &  \multicolumn{3}{c}{Sampling-based~\cite{penicka22RALsbmintimequad}} & \multicolumn{2}{c}{\textbf{RL (ours)}} \\ 
\cmidrule(lr){3-5} \cmidrule(lr){6-7}  \cmidrule(lr){8-9} \cmidrule(lr){10-12}  \cmidrule(lr){13-14}                          &                            & c. time[s] & \multicolumn{1}{c}{$T_{a}$[s]} & $T_{b}$[s] &  c. time[s] & $T_{b}$[s] & c. time[s] & $T_{b}$[s] & c. time[s] & \multicolumn{1}{c}{$T_{a}$[s]} & $T_{b}$[s] & i. time[s] & $T_{b}$[s] \\ 
     \midrule 
\multirow{4}{*}{Forest} &0 &0.41 &4.67$\pm$0.63 &3.86 &17.52 &1.60 &70.23 &\textbf{0.95} &14.92 &1.10$\pm$0.13 &0.96 & 0.00042 &0.98 \\ 
&1 &0.30 &3.43$\pm$0.00 &3.43 &9.34 &1.40 &67.40 &\textbf{0.96} &2.85 &0.97$\pm$0.00 &\textbf{0.96} & 0.00042 &1.00 \\ 
&2 &0.30 &3.74$\pm$0.93 &3.22 &3.12 &1.40 &65.49 &\textbf{0.95} &8.03 &0.98$\pm$0.01 &0.96 & 0.00042 &1.00 \\ 
&3 &1.47 &7.20$\pm$1.17 &5.25 &41.90 &1.80 &- &- &135.83 &1.50$\pm$0.17 &1.30 & 0.00052 &\textbf{1.28} \\ 
\midrule 
\multirow{4}{*}{Office} &0 &1.36 &8.64$\pm$1.03 &7.34 &28.35 &2.60 &- &- &139.19 &2.38$\pm$0.28 &1.93 & 0.00040 &\textbf{1.62} \\ 
&1 &0.65 &7.50$\pm$0.44 &6.49 &100.16 &2.20 &- &- &103.64 &1.74$\pm$0.06 &1.69 & 0.00043 &\textbf{1.64} \\ 
&2 &0.89 &9.01$\pm$0.77 &6.38 &47.08 &2.20 &- &- &155.23 &2.20$\pm$0.13 &1.93 & 0.00045 &\textbf{1.56} \\ 
&3 &0.47 &5.26$\pm$0.35 &5.14 &48.02 &2.00 &- &- &223.64 &1.81$\pm$0.11 &1.58 & 0.00040 &\textbf{1.40} \\ 
\midrule 
\multirow{4}{*}{Racing} &0 &1.98 &6.56$\pm$0.66 &5.79 &- &- &- &- &365.91 &1.61$\pm$0.29 &1.34 & 0.00039 &\textbf{1.20} \\ 
&1 &2.06 &6.21$\pm$0.88 &5.13 &- &- &- &- &428.02 &1.63$\pm$0.15 &1.36 & 0.00042 &\textbf{1.20} \\ 
&2 &1.87 &6.26$\pm$0.42 &4.94 &- &- &- &- &138.17 &1.45$\pm$0.12 &\textbf{1.37} & 0.00039 &1.38 \\ 
&3 &1.91 &5.27$\pm$0.69 &4.73 &- &- &- &- &604.55 &2.14$\pm$0.74 &1.57 & 0.00039 &\textbf{1.34}  \\ 
\midrule 
\multirow{1}{*}{Racing MW} &0 &8.12 &27.54$\pm$0.62 &26.98 &- &- &- &- &734.65 &7.10$\pm$0.06 &7.01 & 0.00037 &\textbf{6.92} \\ 
\bottomrule 
   \end{tabular} 
   } 
   \vspace{-2.0em}
\end{table*}

Table~\ref{tab:single_target} shows that the polynomial method has the highest trajectory duration while being the fastest in computation. 
In contrast, the search-based method generates substantially faster trajectories; however, it requires longer computation times.
The search-based method is severely limited by the state space discretization, which limits finding truly time-optimal continuous-space trajectories. 
Furthermore, the method only returns valid solutions in the 2D Forest and Office environments, while it failed to find valid solutions for the 3D Racing environment and multi-waypoint scenarios.
The optimization-based method can find the time-optimal trajectories only for the three simple Forest test cases.
It fails in the other environments due to the introduced non-convex collision avoidance constraints.
The sampling-based method is capable of finding high-quality solutions for all considered test cases; however, the computational time is comparable to~\cite{Liu_search_based_LQMTC,foehn2020cpc}.

Finally, our learning-based method can achieve on par, mostly better, solutions compared to the conventional methods.
In test cases where the baselines slightly outperform our approach, the difference is within, or close to, the time precision $d_{t}=\SI{0.02}{\second}$ of the RL policy.
We observe that the performance margin of our approach compared to the baselines increases with increasing environment complexity.

In simplistic environments, our learning-based method has lower performance than the best baseline.
This is due to the trade-off between flying safe and flying fast. 
As a result, our approach opts for safer, however slower, actions in cases where a time-optimal trajectory almost touches the obstacles.
This effect is difficult to be modeled when using the optimization-based or the sampling-based method. 
The risk-awareness property of our neural network policy plays an essential role in achieving a high success rate in the presence of a model mismatch. 

In more complex Office and Racing environments, our approach finds significantly faster trajectories than all baselines. 
This is because the sampling-based method uses a hierarchical approach where a point-mass trajectory guides the final trajectory for the quadrotor. 
It leads to rather bang-bang body rate behavior, which is favorable for the simple Forest scenarios. 
In contrast, for the Office scenarios, smooth body rates produced by RL are favorable.
Finally, the inference time of the neural network policy is on average less than one millisecond, which allows fast online adaptation. 

\subsection{Success Rate of Minimum-time Flight in High Fidelity Simulation}

To validate the success rate of navigating through given waypoints in a cluttered environment, we utilize a high-fidelity simulation which is based on Bade-Element-Momentum (BEM) theory~\cite{NeuroBEM}. 
Compared to the simple simulation, the BEM simulation can accurately model lift and drag produced by each rotor from the current ego-motion of the platform and the individual rotor speeds.
Our proposed approach is compared with trajectories planned using the sampling-based method and tracked at \SI{100}{\hertz} using Model Predictive Control (MPC)~\cite{agilicious} that outputs the same thrust and body rates commands as the RL policy.
The output of the MPC is then tracked using the same low-level BetaFlight controller as used for tracking the policy actions.
The sampling-based method is the only other method capable of computing collision-free trajectories for all test cases.
We use MPC for the trajectory tracking, since it has been shown to successfully track truly time-optimal trajectories~\cite{foehn2020cpc}.
The MPC has been tuned to maximize position tracking performance to stick to the planned trajectory collision-free positions and thus avoid obstacles.
The thrust limit of the considered platform is increased from the $f_{max}=\SI{7}{\newton}$ used for planning and learning the policies, to $f_{max}=\SI{8.5}{\newton}$ to have control margins for the MPC.
Furthermore, the BEM simulation uses, in contrast to the simple simulation, both the motor dynamics and the aerodynamics.

Table~\ref{tab:success} shows the comparison of the success rate, measured over 30 runs per test case.
A successful run is defined if the quadrotor passes all waypoints within given tolerance $r_{tol}$ and avoids all obstacles.
The obstacle tolerance $d_{c}$ is decreased by the ESDF precision to remove the influence of the discretized ESDF map representation.
We also show the average $T_a$ and best $T_b$ duration of all tested flights for both methods.

\begin{table}[t] 
   \centering 
   \footnotesize 
   {\renewcommand{\tabcolsep}{3.0pt} 
   \caption{Comparison of success rates of navigating through waypoints while avoiding obstacles.\label{tab:success}} 
   \vspace{-1em} 
   \begin{tabular}{lcrrrrrr}
     \toprule 
    \multirow{2}{*}{Env.} & \multirow{2}{*}{\begin{minipage}{5mm}Test\\case\end{minipage}} & \multicolumn{3}{c}{SB~\cite{penicka22RALsbmintimequad} + MPC~\cite{foehn2020cpc}} & \multicolumn{3}{c}{\textbf{RL (ours)}} \\ 
\cmidrule(lr){3-5} \cmidrule(lr){6-8}                            &                            & success[\%] & $T_{a}$[s] & $T_{b}$[s] & success[\%]  & $T_{a}$[s] & $T_{b}$[s]\\ 
     \midrule 
\multirow{4}{*}{Forest} & 0 & 25 & \textbf{1.22} & \textbf{1.13} & \textbf{100} & 1.23 & 1.21\\ 
& 1 & 0 & \textbf{-} & - & \textbf{100} & \textbf{1.21} & \textbf{1.18}\\ 
& 2 & 27 & \textbf{1.14} & \textbf{1.11} & \textbf{100} & 1.20 & 1.18\\ 
& 3 & 16 & 1.70 & \textbf{1.42} & \textbf{100} & \textbf{1.57} & 1.56\\ 
\midrule 
\multirow{4}{*}{Office} & 0 & 41 & 2.38 & 2.12 & \textbf{100} & \textbf{1.91} & \textbf{1.89}\\ 
& 1 & 28 & 1.86 & \textbf{1.78} & \textbf{100} & \textbf{1.84} & 1.82\\ 
& 2 & 56 & 2.29 & 1.97 & \textbf{100} & \textbf{1.74} & \textbf{1.72}\\ 
& 3 & 70 & 2.16 & \textbf{1.70} & \textbf{100} & \textbf{1.99} & 1.96\\ 
\midrule 
\multirow{4}{*}{Racing} & 0 & 57 & 1.61 & 1.46 & \textbf{100} & \textbf{1.41} & \textbf{1.39}\\ 
& 1 & 51 & 1.64 & 1.45 & \textbf{100} & \textbf{1.47} & \textbf{1.44}\\ 
& 2 & 76 & 1.72 & 1.51 & \textbf{100} & \textbf{1.51} & \textbf{1.49}\\ 
& 3 & 54 & 1.80 & 1.62 & \textbf{100} & \textbf{1.46} & \textbf{1.43}\\ 
\midrule 
\multirow{1}{*}{Racing MW} & 0 & 25 & \textbf{7.22} & \textbf{7.17} & \textbf{100} & \textbf{7.22} & 7.18\\ 
\bottomrule 
   \end{tabular} 
   } 
   \vspace{-1.0em}
\end{table}

Table~\ref{tab:success} indicates that our learned policy can be transferred to a different simulator without fine-tuning. 
The policy successfully navigates the quadrotor through all environments without collisions while achieving faster flight trajectories. 
By contrast, the sampling-based method combined with MPC has significantly lower success rates.
%
This is because the MPC struggles to handle disturbances during high-speed flight.
When the platform is at its actuation limit, the slightest deviation from the pre-planned trajectory results in a suboptimal flight path,
and eventually catastrophic crashes due to the presence of obstacles. 
%

%
%
%

\subsection{Real-world Validation}
We validate our policy in the real world, where the quadrotor has to navigate through the Slalom environment.
Figure~\ref{fig:illustration} shows a successful deployment of the policy. 
The used platform is based on the open-hardware and open-source Agilicious quadrotor framework~\cite{agilicious}.
We use the BetaFlight controller to track the commanded collective thrust and body rates.
We conducted our experiment in the world’s largest indoor drone-testing arena (30 × 30 × 8 \SI{}{\cubic\metre}) equipped with a motion capture system with $\SI{400}{\hertz}$ operation frequency.
The attached video shows the flight with thrust limits of $f_{max}=\SI{4}{\newton}$ and $f_{max}=\SI{7}{\newton}$ where in the second case the maximal velocity reached~\SI{42}{\kilo\meter\per\hour}.
Our policy achieved a flight duration of \SI{7.68}{\second} in simulation, 
while the duration of the real flight was \SI{7.90}{\second}, when using the thrust limit of $f_{max}=\SI{7}{\newton}$. 


\section{Conclusions\label{sec:conclusion}}


This paper introduced a novel method that combines deep reinforcement learning and classical topological path planning to train robust neural network controllers for minimum-time quadrotor flight in cluttered environments.
We showed that the proposed method outperforms existing state-of-the-art approaches in the majority of the test cases, with improved trajectory quality of up to 19\%.
More importantly, we showed that the trained neural network controller can adapt online to counteract disturbances and model mismatches, and thus fly robustly.
The presented method achieves 100\% success rate of flying minimum-time policies without collision, while  approaches relying on the traditional planning and control pipeline achieve only 40\%.
Our findings suggest that model-free deep RL is a promising method for addressing challenging tasks in agile flight, such as dynamic obstacle avoidance or vision-based minimum-time flight in cluttered environments. 


\section*{Acknowledgment}
The authors would like to thank Leonard Bauersfeld for helping with the multimedia material.


\balance

\bibliographystyle{IEEEtran}
\bibliography{main}



\end{document}